%% file: HigherCriticism.tex
\documentclass[preprint]{imsart}

\RequirePackage{amsthm,amsmath,amsfonts,amssymb}
\RequirePackage[authoryear]{natbib}
\RequirePackage[colorlinks,citecolor=blue,urlcolor=blue]{hyperref}
\RequirePackage{graphicx}

\startlocaldefs

\usepackage{graphicx}
\usepackage{psfrag,epsf}

\usepackage{enumerate}
\usepackage[authoryear]{natbib}

\usepackage{amsmath}

\usepackage[utf8]{inputenc} 
\usepackage{url}
\usepackage{mathptmx} 
\usepackage{times} 
\usepackage[export]{adjustbox}
\usepackage{tikz}

\usepackage{lipsum}
\usepackage{algpseudocode}

\usepackage{pgfplots}
\pgfplotsset{compat=1.16}
\usepackage{pstool}
\usetikzlibrary{arrows,shapes,math,patterns}
\usepackage{epic}
\usepackage{epstopdf}

\input{macros}
\endlocaldefs

\usepackage[]{algorithm}

\begin{document}

\def\noworks{11,050}
\definecolor{mygrey}{RGB}{229,229,229}
\definecolor{mygrey2}{RGB}{127,127,127}
\definecolor{mygrey3}{RGB}{240,240,240}
\pgfplotsset{
     axis background/.style={fill=mygrey},
    tick style=mygrey2,
    tick label style=mygrey2,
    grid=both,
    xtick pos=left,
    ytick pos=left,
    tick style={
        major grid style={style=white,line width=1pt},minor grid style=mygrey3,
        tick align=outside,
    },
    minor tick num=1,
}

\begin{frontmatter}
\title{Higher Criticism for Discriminating Word-Frequency Tables and Authorship Attribution}
\runtitle{Higher Criticism for Word-Frequency Tables}

\begin{aug}
\author[A]{\fnms{Alon} \snm{Kipnis}\ead[label=e1]{kipnisal@stanford.edu}},

\address[A]{Department of
Statistics,
Stanford University,
\printead{e1}}

\end{aug}


\begin{abstract}
We adapt the Higher Criticism (HC) goodness-of-fit test to measure the closeness between word-frequency tables. We apply this measure to authorship attribution challenges, where the goal is to identify the author of a document using other documents whose authorship is known. The method is simple yet performs well without handcrafting and tuning; reporting accuracy at the state of the art level in various current challenges. %
As an inherent side effect, the HC calculation identifies a subset of discriminating words. %
In practice, the identified words have low variance across documents belonging to a corpus of homogeneous authorship. 
We conclude that in comparing the similarity of a new document and a corpus of a single author, 
HC is mostly affected by words characteristic of the author and is relatively unaffected by topic structure. 
\end{abstract}

\end{frontmatter}

\section{Introduction}
\label{sec:intro}

The unprecedented abundance and availability of text data in our age generates many  
{\it authorship attribution problems} of the following form. We obtain a new document of unknown authorship; we would like to determine its author.
We also have data: several corpora of documents, each of homogeneous authorship. We believe the unknown author of the new document is represented among our corpora and we wish to attribute authorship to the new document based on our data.
Existing approaches for such problems usually construct a set of handcrafted features to discriminate between potential candidate authors \citep{MostellerWallace,holmes1985analysis,EfronThisted1986,tilahun2012dating,zheng2006framework,juola2008authorship,Glickman2019Data}. 
Typically, these features originate from linguistic heuristics, such as rate of use of certain words and length of sentences, 
and are often first constructed by trial and error, or based on domain expertise or historical tradition.   \par
While this process sometimes achieves convincing and widely accepted results, it is not automatic. 
The discriminating features and test statistics are crafted for each specific problem, 
and it is unclear whether these features or tuned parameters can be reused in other problem domains. 
A famous example that demonstrates these limitations is Mosteller and Wallace's work on authorship in the Federalist Papers \citep{MostellerWallace}, 
a collection of articles explaining the nascent US constitution -- written between October 1787 and September 1788 by Alexander Hamilton, James Madison, and John Jay. All articles were published under a single pseudonym, regardless of actual authorship. The identities of the three authors, as well as the specifics of who wrote each article, were revealed or claimed in subsequent years. Among the first 77 articles, historical sources agree that Jay wrote 5 articles, Hamilton wrote 43, Madison wrote 14, 3 articles were written jointly by all three,
while the authorship of the remaining 12 is disputed between Hamilton and Madison. Mosteller and Wallace determined that all 12 disputed papers are the sole work of Madison. Their process involves two major steps: 
\begin{itemize}
    \item[(i)] Identifying discriminating words, i.e., words whose frequencies in known Hamilton texts are different from those of Madison's.  
    \item[(ii)] Combining frequencies of these words in articles of known authorship and disputed ones to a single test statistic. 
\end{itemize}
The specifics of these steps are described in \citep{MostellerWallace} and \citep{mosteller2012applied}. In a nutshell, step (i) relied on linguistic assumptions for considering an initial list of 176 ``non-contextual'' words and some selection procedure to reduce this list. Step (ii) involved various Bayesian modeling decisions as well as some heuristics for estimating the parameters of these models. In particular, it appears that the overall procedure obtained from (i) and (ii) cannot be applied to other authorship challenges without significant modifications. Indeed, we are unaware of other authorship studies that have applied word elimination processes or modeling choices akin to \citep{MostellerWallace}. 
\par
In this paper, we describe a technique of authorship attribution that can be used ``out-of-the-box''. When applied to standard authorship challenges, it performs about as well as other approaches but without handcrafting and tuning. 

Our technique relies on a relatively simple statistical tool: it uses the Donoho-Jin-Tukey Higher Criticism (HC) statistic as a measure of closeness between word-frequency tables (viz. bag-of-words) \citep{donoho2004higher}. We select the likely author using proximity under this measure. 
The resulting procedure is automatic in the sense that it does not require prior screening for discriminating words or features.
In fact, it inherently identifies a set of likely discriminating features during the calculation of the HC statistic. 
As we show below, the set thus identified often corresponds to words whose counts exhibit low variance across documents within a corpus of homogeneous authorship. 
Consequently, for comparing a new document with the corpus of a known author, this proximity measure seems most affected by the words characteristic of that author and is relatively unaffected by the topic structure of the text. 
\\

The basic tool we develop in this paper is a technique to discriminate between two word-frequency tables which might both be sampled from the same source frequencies, or else perhaps not. Here `word frequencies' extend in an obvious fashion to n-gram or to frequencies of other features of the text that can be summarized as entries in a frequency table. Aside from the authorship attribution problem, n-gram frequency tables have been proven to be a useful summary of textual data more broadly in information retrieval and linguistics \citep{manning2010introduction, MargaretAiroldi2016}. It is straightforward to adapt the approach described here to other text classification problems besides authorship attribution. \\

\begin{figure}
    \centering
    \includegraphics[width=2.5in]{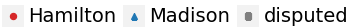}
    
    \includegraphics[width=2.6in]{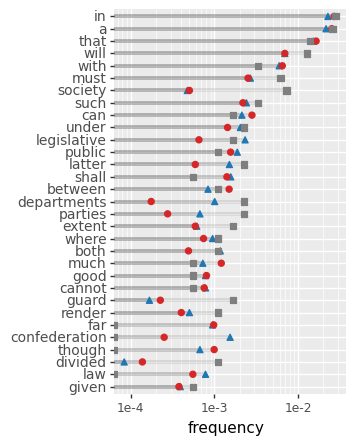}
    \includegraphics[width=2.6in]{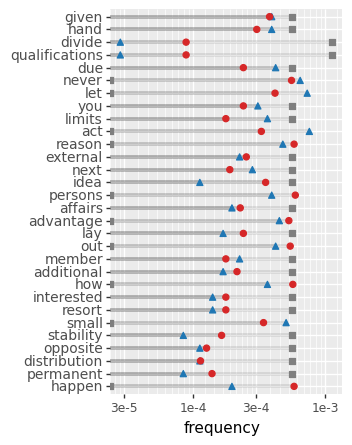}

    \caption{Three word-frequency tables in the Federalist Papers. \newtext{Word frequencies of a random sample of
    60 words out of the 500 most common ones in one of the disputed articles (gray), the corpus of known Hamilton articles (blue) and the corpus of known Madison articles (red) articles out of the first 77 Federalist Papers. We attribute the disputed article by measuring the global discrepancy between its word frequencies to each corpus of known authorship.}
    \label{fig:the_challenge}}
\end{figure}

\subsection{Discriminating Word-Frequency Tables}
Figure~\ref{fig:the_challenge} illustrates word frequencies in the first 77 Federalist Papers studied in \citep{MostellerWallace}, divided into three corpora: the circles and triangles represent the frequencies of words in Hamilton's and Madison's known documents, respectively. The squares represent word frequencies from one of the 12 disputed papers. Our goal is to determine which of the two word-frequency tables of known authors best resembles the word-frequency table of the unknown author's document. 
\par
Standard approaches to this problem include two-sample tests for homogeneity of discrete multivariate data such as power divergence tests 
\citep{bishop2007discrete,read2012goodness}. 
It has long been observed, however, that these tests are not optimal in the high-dimensional setting where the number of entries in the table is large compared to the size of the sample \citep{hoeffding1965asymptotically,balakrishnan2018hypothesis} or when the frequencies are imbalanced \citep{arias2011global}. 
This high-dimensional setting is the typical situation in word frequency tables representing natural text. 
Moreover, the form of alternatives considered in analyzing and developing classical tests of homogeneity is quite general, whereas the important differences in word frequencies between authors may be concentrated on a sparse subset. Namely, relatively few words, out of possibly thousands, may indicate a change of authorship. 
Consequently, a test that adapts well to sparsity seems promising in this application. In addition to the rareness of discriminating words, the evidence that each such word provides is weak; no single word serves as a decisive discriminating feature. 
To summarize, we are facing the problem of detecting a \emph{rare} change in the distribution of a large set of possibly \emph{weak} features. HC has long been known to detect signals of a rare/weak nature \citep{donoho2004higher,arias2011global,mukherjee2015hypothesis,donoho2015special,li2015higher, jin2016rare}. This motivates us to adapt HC to our purpose of detecting changes between word frequency tables. 

\subsection{Binomial Allocation Model}
We think about a document as an ordered list of words.
Given a vocabulary $W$, the word-frequency table associated with the document $D$ is denoted by $\{N(w|D),\, w \in W\}$, where $N(w|D)$ records the total number of occurrences the word $w$ in $D$. \\

\newtext{Consider two documents $D_1$ and $D_2$. For each occurrence of a word $w\in W$ in either document, place in a database the labelling pair $(w,l)$ where $w$ denotes the word and $l$ the label "1" or "2" according to which document contains that occurrence. %
Suppose that, under the null hypothesis, different occurrences are independent and that each is equally-likely to originate from "1" (respectively "2"), only accounting for the relative size of $D_1$ compared to $D_2$ minus occurrences of $w$. Equivalently, occurrences of $w$ are obtained by sprinkling the records in the database with the labels removed across the remaining locations in the large document obtained by concatenating $D_1$ and $D_2$.}
In this case, 
\[
N(w|D_1) \sim \Bin(n,p),
\]
where\footnote{\newtext{The binomial model would not be correct unless we omit occurrences of $w$ when considering the relative size of $D_1$ to define $p$.}}
\begin{align}
n = N(w|D_1) +  N(w|D_2), \quad 
p = \frac{\sum_{w'\in W, \, w'\neq w} N(w'|D_1) }{\sum_{w'\in W, \, w'\neq w} \left(N(w'|D_1)+N(w'|D_2)\right) }. \label{eq:p_def}
\end{align}
The hypothesis test
\begin{align}
    \label{eq:binomial_test}
\begin{cases}
H_0~ &:~~ N(w|D_1) \sim \Bin(n,p) \\
H_1~ &:~~ N(w|D_1) \sim \text{not }\Bin(n,p)
\end{cases}
\end{align}
has an exact P-value under the null hypothesis\footnote{For example, see the \texttt{R} function \texttt{binom.test}.}, roughly, 
\begin{align}
\pi(w|D_1,D_2) \equiv \Prob\left(\left|\Bin(n,p) - n p \right| \geq \left|N(w|D_1) - n p \right| \right).
\end{align}
Applying this test word-by-word, we obtain a large number of P-values $\{ \pi(w|D_1,D_2)\}_{w\in W}$. \newtext{We apply the Higher-Criticism (HC) statistic to these P-values, obtaining a global hypothesis test against the null hypothesis that all obey the binomial allocation model outlined earlier.}

\subsection{The Higher Criticism}
The HC of the P-values $\{p_i\}_{i=1}^N$ is defined as
\begin{align}
    \label{eq:HC}
\HC^* \equiv 
\underset{{1\leq i \leq \gamma_0N}}{\max} \sqrt{N} \frac{i/N - p_{(i)}}{\sqrt{\frac{i}{N}\left(1-\frac{i}{N}\right)}},
\end{align}
where $p_{(i)}$ is the $i$-th P-value among $\{ p_i,\, i=1,\ldots,N \}$ and $\gamma_0$ is a tunable parameter.\footnote{ $\HC^*$ and $\HC^\dagger$ appear to be insensitive to the choice of $\gamma_0$ provided $|W|$ is large enough. Our experience shows that the choice $\gamma_0 \in (0.2,0.35)$ provides good results in moderate sample sizes where $|W| > 100$.} The $\HC$ test takes a large batch of P-values and returns a single number, indicating the global significance of the body of P-values \citep{donoho2004higher,donoho2008higher}. 
\newtext{The idea behind HC goes back to Tukey, who proposed a way to measure the global significance of many level-$\alpha$ independent tests by considering the difference between the standardized z-scores of the observed fraction of tests that are significant to their expected fraction under the joint null. Donoho and Jin proposed to use $\HC^*$, the maximized z-scores over the range of significance levels $0 \leq \alpha \leq \gamma_0$, as a global test against the joint null \citep{donoho2004higher}. Their proposal have shown to be effective in resolving several challenging testing problems \citep{cai2007estimation,jager2007goodness,delaigle2009higher,ingster2010detection,hall2010innovated, arias2011global, tony2011optimal, arias2015sparse,mukherjee2015hypothesis}.}

\newtext{In our adaptation of the HC test to word-frequency tables, we define the HC-discrepancy $\dist_{\HC}(D_1,D_2)$ of documents $D_1$ and $D_2$ using the following variant of the HC test statistic:}
\begin{align}
    \label{eq:HC_dagger}
 \dist_{\HC}(D_1,D_2)  \equiv \HC^\dagger \equiv \underset{\substack{1 \leq i \leq \gamma_0 N \\ 1/N \leq \pi_{(i)}}}{\mathrm{max}} \sqrt{N} \frac{i/N - \pi_{(i)}}{\sqrt{\frac{i}{N}\left(1-\frac{i}{N}\right)}},
\end{align}
\newtext{where $\pi_{(i)}$ is the $i$th P-value among $\{ \pi(w|D_1,D_2) \}_{w\in W}$ and $N = |W|$ is the size of the vocabulary $W$ (note the symmetry $\dist(D_1,D_2)=\dist(D_2,D_1)$). The statistic $\HC^\dagger$ is based on a proposal of \citep{donoho2004higher} for improving the numerical stability of $\HC^*$. Our experience shows that $\HC^\dagger$ performs slightly better than $\HC^*$ in authorship challenges; see the results in Table~\ref{tab:Gutenberg_results}.}

The procedure for obtaining the HC-discrepancy of two documents is summarized in Algorithm~\ref{alg:1}. We extend this procedure to measure the discrepancy between a document and a corpus by thinking about this corpus as the concatenation of all documents within it. \newtext{It is generally challenging to use the HC-discrepancy to conduct a level-$\alpha$ test against a null hypothesis of the form ``the given document and corpus are of the same author''. We consider such test \citep{kipnisPAN2020} in the context of the authorship verification challenge of \citep{kestemont2020overview}. Our experience shows that this test requires large amounts of calibration data, hence it is impractical in most real-world authorship challenges. Instead, in this paper we focus on the authorship attribution problem, i.e., we associate a document of unknown authorship to one author among several candidates.}

\begin{figure}
\begin{center}
\begin{tikzpicture}
\node at (0,0) {\includegraphics[scale = 0.7]{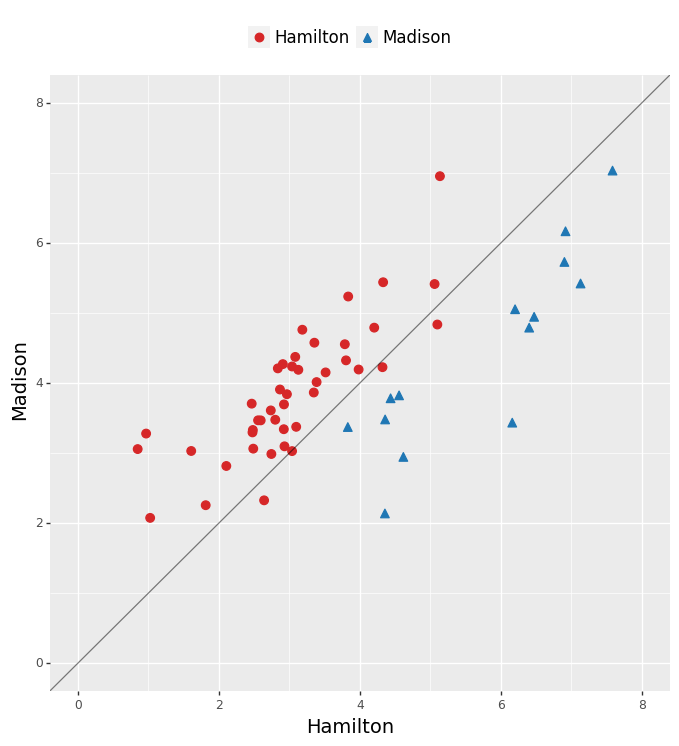}};
\node[align=center] at (3,-3) {similar to \\ Madison
};
\node[align=center] at (-3,3) {similar to \\ Hamilton
};
\end{tikzpicture}
\end{center}
\caption{
\newtext{Authorship in the Federalist Papers. Each point indicates HC-discrepancy of a document by Madison (red) or Hamilton (blue) with respect to Hamilton's corpus of 43 papers ($x$-axis) and Madison's corpus of 14 papers ($y$-axis) (in comparing a document to the corpus of its own author, this document is left out of the corpus). The diagonal line $y=x$ is indicated. With a few exceptions, Madison articles lie below the diagonal, while Hamilton's lie above the diagonal.}
}
    \label{fig:sep}
\end{figure}

\newtext{Figure~\ref{fig:sep} illustrates HC-discrepancies of Hamilton's corpus versus Madison's in the Federalist Papers: each point indicates the HC-discrepancy of one document compared to either corpus. This figure suggests that it is possible to correctly attribute authorship with high accuracy by using HC as an index of discrepancy.} 

We emphasize that we are not relying on an assumption that the underlying generative model of binomial word allocation be exactly true; there may well be departures such as correlations and overdispersion. Again, HC is here being used as an 
an index of discrepancy. 


\subsection{The HC Threshold}
Associated with the HC statistic \eqref{eq:HC} is the \emph{HC threshold} $t_{\HC}$, defined as
\begin{align}
    \label{eq:HC_thresh}
t_{\HC} \equiv \pi_{(i^*)}, \quad i^* \equiv \underset{\substack{1 \leq i \leq \gamma_0 N \\ 1/N \leq \pi_{(i)}}}{\argmax}  \frac{i/N - \pi_{(i)}}{\sqrt{\frac{i}{N}\left(1-\frac{i}{N}\right)}}.
\end{align}
Roughly speaking, the HC statistic describes the maximal deviation of the collection of P-values $\{\pi(w|D_1,D_2),\, w \in W\}$ from the uniform distribution over $(0,1)$. This deviation is mostly affected by P-values that fall below $t_{\HC}$. 
In \citep{donoho2009feature}, it was shown that the HC threshold leads to an optimal feature selection procedure in some classification settings. In  \citep{jin2016influential}, the HC threshold is applied for  selecting features in a specific clustering setting. 
In our context, we propose to use the HC threshold to identify words distinguishing between the two word-frequency tables, and, consequently, between the author associated with each of these tables.  
\\

The procedure outlined above for measuring the discrepancy of two word-frequency tables and obtaining a set of discriminating words is summarized in Algorithm~\ref{alg:1}.

\begin{algorithm}
\textbf{Input:} Two word-frequency tables $\{N(w|D_1),\,w\in W\} $ and $\{N(w|D_2),\,w\in W\}$. 
\begin{algorithmic}
\Procedure{HC-discrepancy}{} 
\State $n_1 \leftarrow  \sum_{w\in W} N(w|D_1)$;
\State $n_2 \leftarrow  \sum_{w\in W} N(w|D_2)$;
\For {$w \in W$}
\State $x \leftarrow N(w|D_1)$;
\State $n_w \leftarrow N(w|D_1) + N(w|D_2)$;
\State $ p_w \leftarrow \frac{ n_1-x  } {
n_1+n_2-n_w}$;
\State $\pi(w|D_1,D_2)$ $\leftarrow$ binomial test $(x,n_w,p_w)$;
\EndFor
\State $N$ $\leftarrow$ $|W|$; 
\State $\left( \pi_{(1)},\ldots,\pi_{(N)} \right)$ $\leftarrow$ $\mathrm{Sort}( \{ \pi(w|D_1,D_2) \}_{w\in W})$; 
\State $z_i \leftarrow  \sqrt{N} \frac{i/N - \pi_{(i)}}{\sqrt{\frac{i}{N}\left(1- \frac{i}{N} \right) }}$, $i=1,\ldots,N$;
\State $i_{\min} \leftarrow \argmin \{ i,\, 1/N \leq \pi_{(i)} \}$;
\State $i^* \leftarrow \argmax \{ z_i,\, i_{\min} \leq i \leq \gamma_0 N\}$; 
\State $\HC^\dagger \leftarrow z_{i^*}$; 
\State $W^\dagger$ $\leftarrow$ $\left\{w \in W \,:\, \pi(w|D_1,D_2) \leq \pi_{(i^*)} \right\}$;\\
~~~~\Return{$\HC^\dagger$, $W^\dagger$}
\EndProcedure
\end{algorithmic}
\caption{HC-discrepancy of word-frequency tables \label{alg:1}} 
\end{algorithm}

\subsection{Analyzing Ingredients for Success}
Textual data serves as a channel to deliver information in multiple contexts. It is therefore challenging, or perhaps impossible, to provide a comprehensive theory for the performance of the HC-discrepancy that covers all authorship attribution scenarios. 
Instead, in order to understand the empirical success of our test in attributing authorship, we analyze the properties of the words that fall below the HC threshold since these are the words affecting the HC-discrepancy most. 
For this purpose, we apply variance-stabilizing transformations to word-frequency tables and compare the variance associated with the same word across a corpus of homogeneous authorship to the P-value associated with this word under a binomial allocation model. 
By examining a large number of pairs of authors, we discover that words having the most influence on the value of the HC statistic are associated with small variances across documents in each author's corpus. 
This finding shows that the HC-discrepancy is not heavily affected by the topic structure of the text. It seems that words 
contributing to the test statistic are characteristic of the author's style rather than the characteristic of a particular topic.

\subsection{Related Works}
\newtext{
The problem of testing hypothesis based on frequency or contingency tables dates back, at least, to Pearson \citep{pearson1900x}, whose chi-square test is still the standard choice in this problem. We refer to the classical book \citep{bishop2007discrete} as an introduction to the topic. 
The one-sample version of the problem, in which the observed frequencies are replaced by the true underlying frequencies in one of the tables, appears under the names: testing multinomial, goodness-of-fit with categorical data, and, in computer science, distribution identity testing. }
In accordance with modern challenges in data analysis, there is much recent interest in the high dimensional version of this problem in which the number of samples is small compared to the size of the vocabulary, the number of categories, or the support of the distribution. See 
\citep{balakrishnan2019hypothesis} and the related review paper \citep{balakrishnan2018hypothesis}. Furthermore, the work of \citep{DonohoKipnis2020} studies the asymptotic properties of HC under rare and weak perturbations of the categories. We note that choices for combining the binomial allocation P-values other than \eqref{eq:HC_dagger} may be preferable in some cases; see  \citep{li2015higher} for a discussion. Our choice of HC here is largely motivated by its well-understood feature selection mechanism, the HC threshold \citep{donoho2009feature}.

\par
Authorship studies in the statistical literature include, most notably, the case of the Federalist Papers \citep{MostellerWallace, mosteller2012applied}.
The surveys \citep{holmes1985analysis} and \citep{juola2008authorship} provide wide coverage of the topic. 
%
Another line of statistical works concerning authorship first identifies some regularity property of the text and then uses deviations from the regular behavior to attribute or refute authorship \citep{Wake1957,cox1959discriminatory,Sichel1974}. This practice was also adopted by 
Efron and Thisted \citep{efron1976estimating, EfronThisted1986}, who applied their estimator of the number of unseen species to determine if the number of novel words in a disputed text matches the degree of novelty had Shakespeare been the author of the text.
\citep{ross2019tracking} addressed the possibility that the style of an author changes over time, and suggested ways to account for this change in authorship studies. \citep{tilahun2012dating} tracked changes in word-frequencies over time to date medieval charters. Very recently, \citep{Glickman2019Data} considered harmonic and melodic features to determine the degree of collaboration in a few famous songs by The Beatles. 

\subsection{Structure of the paper}
The paper is organized as follows: In Section~\ref{sec:testing} we develop a method for attributing the authorship based on the procedure outlined in Algorithm~\ref{alg:1} and applied it in various authorship attribution challenges. In Section \ref{sec:success}, we explain why our method works. Concluding remarks are provided in Section \ref{sec:cocnlusion}. 

\section{Authorship Attribution
\label{sec:testing}}
In this section, we develop a method for text classification and authorship attribution using $\dist_{\HC}$ and evaluate its performance on several datasets. 

Given a document $D$ and a corpus $\Ccal$ not containing $D$, the HC-discrepancy $\dist_{\HC}(D_,\Ccal)$ associated with $D$ and $\Ccal$ provides an index of discrepancy between the document and the corpus.
We suggest using this index of discrepancy to solve the following classification problem: Let $\Ccal_A = \{ D_i,\,i\in I_A\}$ and $\Ccal_B = \{D_i,\,i\in I_B\}$ be two disjoint corpora. 
Upon introducing a new document $D$ that is neither a member of $\Ccal_A$ nor $\Ccal_B$, associate $D$ with one of corpus $A$ or $B$. 
\par
%
%
In order to fix notation, henceforth we identify a corpus $\Ccal$ with the document formed by concatenating all documents in $\Ccal$. 
We also use the notation 
\[
\Ccal_{(D)} \equiv \{ D' \in \Ccal,\, D' \neq D\},
\]
to denote the corpus $\Ccal$ with the document $D$ removed. 

\subsection{Discrepancy between a Document and a Corpus}

Considering a corpus as one large document, we can naturally extend the HC-discrepancy between document-pair to discrepancy between a document $D$ a corpus $\Ccal$. In this case, we set 
\begin{align*}
    \HC_{D|\Ccal} \equiv  \dist_{\HC}(D,\Ccal).
\end{align*}

Figure~\ref{fig:sep} depicts an $x-y$ scatter plot in which each point represents a document in the combined set $\Ccal_{\Ham} \cup \Ccal_{\Mad}$. For $D \in \Ccal_{\Ham}$, 
\[
(x,y) = \left(
\HC_{D|\Ccal_{\Ham}(D)}, \HC_{D|\Ccal_{\Mad}} \right),
\]
while for $D \in \Ccal_{\Mad}$, 
\[
(x,y) = \left( \HC_{D|\Ccal_{\Ham}}, \HC_{D|\Ccal_{\Mad}(D)} \right).
\]
\newtext{It follows from Figure~\ref{fig:sep} that the HC-discrepancy between $D \in \Ccal$ and $\Ccal(D)$ is small compared to the HC-discrepancy between $D$ and the corpus of the other author.} Since points corresponding to documents of opposing authorship are largely separated by the identity line ($y=x$), HC-discrepancy can determine the true author of a new document with high accuracy. \\

\subsection{Rank-based Calibration in Authorship Attribution
\label{sec:rank_calibration}}
Due to the complicated structure of most texts, we do not expect the binomial model underlying our method to be strictly correct; nor do we expect the identity line to be the best discriminator between the two corpora. Instead, we deploy the HC statistic using a rank-based calibration. Consider the rank of the HC-discrepancy of a new document relative to the HC-discrepancy obtained from other documents within a corpus. This rank furnishes a \emph{calibrated} index of discrepancy between the disputed document and the corpus. We assign the document $D$ to whichever corpus gives the smallest normalized rank. We formalize this process using a rank-based testing procedure \citep{d1975nonparametrics}: For each corpus $\Ccal_{\alpha} = \{ D_j,\, j\in I_{\alpha}\}$ and document $D_i$, $i \notin I_\alpha$, consider the extended corpus ${\Ccal}_{\alpha+i} \equiv \left\{ D_j,\, j \in I_{\alpha} \cup \{i\} \right\}$. The null hypothesis $H_{0,\alpha+i}$ states that all scores 
\[
\HC_{{\Ccal}_{\alpha+i} } \equiv \left\{ \HC_{D_j|{\Ccal}_{\alpha}(D_j) } \right\}_{j \in I_{\alpha} \cup \{i\}},\qquad i \notin I_\alpha
\]
are sampled independently from the same continuous distribution over the reals. A P-value with respect to $H_{0,\alpha+i}$ is $1-\hat{r}_{D_i|\Ccal_{\alpha}}$, where 
\begin{equation}
 \hat{r}_{D_i|\Ccal_{\alpha}} \equiv \frac{\mathrm{rank}\left(\HC_{D_i|\Ccal_\alpha}\mid \HC_{{\Ccal}_{\alpha+i}} \right) }{|I_{\alpha}|+1}, \label{eq:Pvals}
\end{equation}
is the rank of $\HC_{D_i|\Ccal_\alpha}$ in the sample $\HC_{{\Ccal}_{\alpha+i}}$.
We consider large values of $\hat{r}_{D_i|\Ccal_{\alpha}}$ to be evidence against the hypothesis that $D_i$ and the other documents in $\Ccal_{\alpha}$ were sampled from the same distribution and hence were written by the same author. 
Consequently, we associate the document $D_i$ to whichever corpus has a smaller $\hat{r}_{D_i|\Ccal_{\alpha}}$. 






\begin{center}
\begin{table}
    \begin{tabular}{|c|c|c|c|c|}
    \hline
         \bf collection & \bf \# authors & \bf \# documents
           & 
     \multicolumn{2}{|c|}{\bf \# words per doc} \\
      & & \bf per author  & (range) & (average) \\
        \hline 
        The Federalist  & 2 &
        \begin{tabular}{cc}
              Hamilton & 53 \\
              Madison & 14 
         \end{tabular}
        & 958 - 3.5k & 2k \\
        \hline 
        \begin{tabular}{cc}
        \noworks~literary works 
        \\
        \end{tabular}
        & 
        488
        & 10-100 (average 23)
        & 10k - 2600k
        & 74k \\ 
      \hline
      \begin{tabular}{cc}
             PAN2018 authorship\\ attribution challenge \\ (problems 1-4)
        \end{tabular}
         & 
         \begin{tabular}{cc}
        $5$, $10$, $15$, $20$ \\ 
        (per problem)      
        \end{tabular}
        & 7 (in training set) & 600 - 1k & 970 \\
    \hline 
    \end{tabular}
    \caption{ Three collections for authorship attribution analysis.
    \label{tab:datasets}}

\end{table}
\end{center}

\begin{figure}
    \centering
\begin{tikzpicture}
\node at (-1,0) {\includegraphics[scale = 0.55]{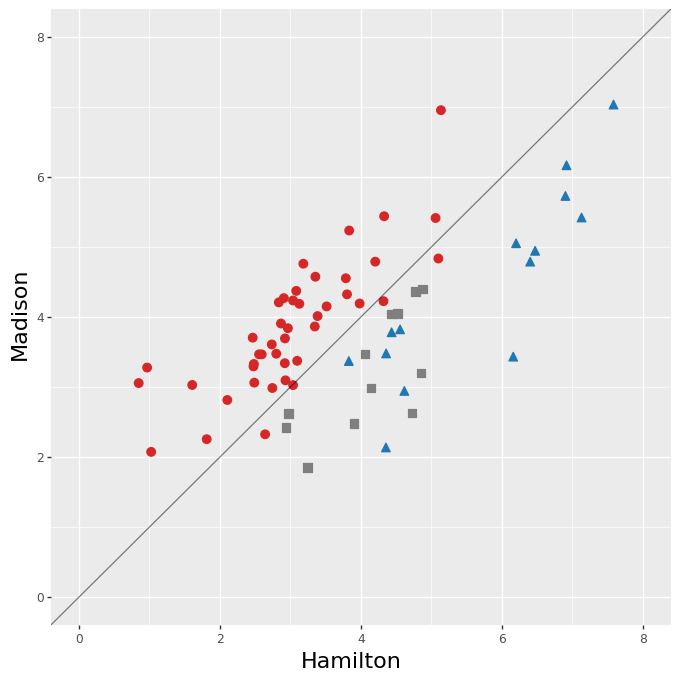}};
\node[align=center] at (2,-3) {similar to \\ Madison
};
\node[align=center] at (-3,2) {similar to \\ Hamilton
};
\node at (6,0) {
\includegraphics[scale = 0.55]{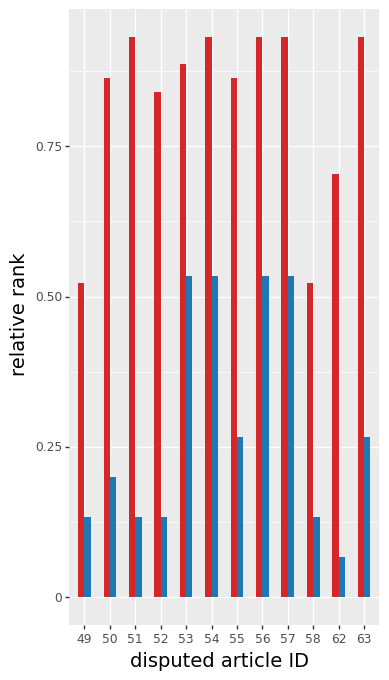}};
\node[fill=white,draw=black] at (2,4.3) {\includegraphics[scale=.7]{Figs/word_freq_table_legend.png}};

\end{tikzpicture}
%


\includegraphics[scale = 0.7]{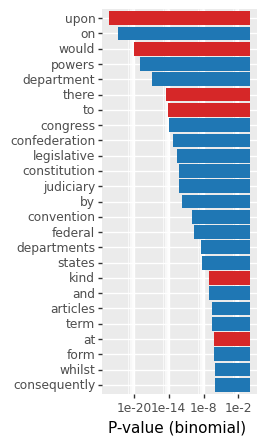}
\includegraphics[scale = 0.7]{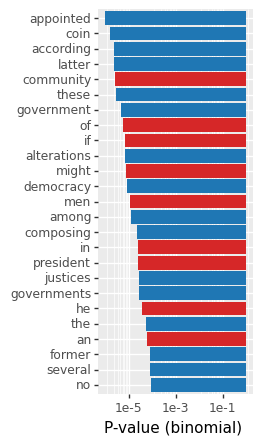}
\includegraphics[scale = 0.7]{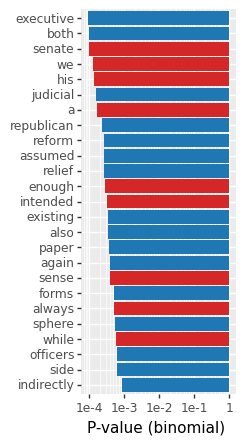}

\caption{ \newtext{
    Authorship in the Federalist Papers. Top-left: 
    HC-discrepancies of each papers with respect to Hamilton's corpus of 43 papers ($x$-axis) and Madison's corpus of 14 papers ($y$-axis). The diagonal line $y=x$ is indicated. 
    This scatterplot is an extended version of Figure~\ref{fig:sep} with additional points corresponding to the 14 disputed papers.}
    Top-right: the relative rank of each disputed article with respect to each corpus (lower rank indicates better similarity). 
    Bottom: Words and their associate P-value in the set $W^\dagger$ returned by $\mathrm{HC-DISCREPANCY}$ when the entire Hamilton's corpus is compared to the entire Madison's corpus. The P-value for each of these words obtained from the test \eqref{eq:binomial_test} falls below the HC threshold. The color of the bar signals the corpus within which the word is more frequent: Red=Hamilton; Blue=Madison. 
    The vocabulary is the union of the set of 1,500 most common words by each of Hamilton and Madison in the Federalist collection with proper names and cardinal numbers removed.
    \label{fig:Federalists_full}
    }
\end{figure}

\subsection{Performance in Authorship Attribution}
We consider the performance of our HC-based method in various authorship attributing challenges. The code and data for challenges in provided in the supplementary material \cite{kipnis2021supp}. 

\subsubsection{The Federalist Papers}
Figure~\ref{fig:Federalists_full} illustrates the HC-discrepancy of each of the 12 disputed Federalist papers (Numbers 49-58, 62, and 63) with respect to Madison's and Hamilton's corpus, respectively.

Also shown in this figure are the normalized ranks  $\hat{p}_{D_i|\Ccal_\alpha}$ of the $i$th disputed paper for $\alpha \in \{ \text{Hamilton}, \text{Madison}\}$. Based on our rank-based calibration, each of the disputed documents seems to be written by Madison rather than Hamilton. 
In testing all Hamilton's corpus against Madison's, \numOfFedWords words whose P-value under the binomial word allocation model is smaller than the HC threshold. 
The bottom of Figure~\ref{fig:Federalists_full} indicates the P-values of $63$ of these words that are also included in the list of ``non-contextual'' words considered in \citep{MostellerWallace}.  

\subsubsection{Large Collection of Potential Authors}
We now assess our procedure to determine authorship from among many authors. We use the Gutenberg Project\footnote{Project {G}utenberg (n.d.), retrieved September 10, 2019, from www.gutenberg.org} to form a collection of texts by 488 authors satisfying our inclusion criterion: At least 10 works with at least 10,000 words. We use a vocabulary consisting of the $N$ most common words in English Google books according to the list \citep{mostCommon}, for $N \in \{250, 1000, 3000\}$. 
For each vocabulary size, we measure the discrepancy of each work and each of the $488$ corpora associated with each author in a 10-fold cross-validation procedure: the entire dataset containing \noworks~works is randomly split into  
10 disjoint subsets. For $i=1,\ldots,10$, all subsets except subset $i$ are used as the training set and accuracy is evaluated for attributing authorship of works in subset $i$. The reported accuracy is averaged over all 10 cases. 
We attribute the work to the author whose corpus attained the smallest HC-discrepancy. \newtext{In this evaluation, we also considered the HC variant $\HC^*$ of \eqref{eq:HC} in addition to $\HC^\dagger$ of \eqref{eq:HC_dagger}}. The average accuracy in this procedure is reported in Table~\ref{tab:Gutenberg_results}
We also used an analogous attribution procedure based on several other discrepancy measures:
\begin{itemize}
    \item \emph{Cosine Discrepancy}. The cosine discrepancy between documents $D_1$ and $D_2$ is defined as
    \[
    \cossim(D_1,D_2) \equiv 1-\cos\left(D_1,D_2\right),
    \]
    \[
    \cos\left(D_1,D_2\right) \equiv \frac{\sum_{w \in W} N(w|D_1)N(w|D_2)}{
    \sqrt{\sum_{w \in W} \left(N(w|D_1)\right)^2} \sqrt{\sum_{w\in W}  \left(N(w|D_2) \right)^2}}.
    \]
    We also considered a nearest neighbor (NN) classifier with $\cossim(D_1,D_2)$ as its underlying metric.
    \item \emph{Power-divergence}. The power-divergence test statistic with a real parameter $\lambda$ is defined as \citep{read2012goodness} 
    \begin{align*}
    & \PD(D_1,D_2) \equiv \\
    &  \sum_{\substack{
    w \in W' \\
    i\in\{1,2\}}}  N(w|D_i)
    \left( \left( \frac{N(w|D_i)}{ \mathrm{T}(w|D_1,D_2)} \right)^\lambda - 1 \right). 
    \end{align*}
    Here $W'$ is the set of words in $W$ such that $N(w|D_1) + N(w|D_2) > 0$, and 
    \[
    \mathrm{T}(w|D_1,D_2) \equiv \alpha_1 N(w|D_1) + (1-\alpha_1) N(w|D_2), 
    \]
    where 
    \[
    \alpha_1 \equiv \frac{\sum_{w\in W}N(w|D_2)}{\sum_{w\in W} N(w|D_1)+N(w|D_2)}.
    \]
    We considered the cases $\lambda=1$ (Pearson's chi-squared test statistic), $\lambda = 2/3$ suggested in \citep{cressie1984multinomial}, and $\lambda\to 0$ corresponding to the likelihood ratio statistic $G^2$. The index of discrepancy corresponding to the power-divergence test statistics is $\PD/(N'-1)$, where $N'$ is the number of words in $W$ such that $N(w|N_1) + N(w|D_2) > 0$. 
\end{itemize}


\begin{table}[t]
    \centering
    \input{table_results}
    \caption{
    Accuracy in determining the authorship of \noworks~ literary works among 488 writers using several statistics (interpreted as indices of discrepancy) and vocabulary sizes. Each $k$-size dictionary consists of the $k$ most frequent English words according to the list in \citep{mostCommon}. Reported accuracy is obtained using a 10-fold cross validation procedure; standard errors are in brackets. }
     \label{tab:Gutenberg_results}
\end{table}
The average accuracy and standard error of each authorship attribution method over difference splits in the Gutenberg authorship challenge are provided in Table~\ref{tab:Gutenberg_results}. \newtext{It follows that HC-discrepancy attains the best accuracy among all the discrepancy methods we tried. Welch's t-test imply that all differences between the accuracies of the HC-discrepancy (based on either $\HC^*$ or $\HC^\dagger$) and the other methods are \emph{significant} at the level $.05$. The difference between $\HC^\dagger$ and $\HC^*$ is significant only for vocabulary size =$3,000$.
}


\subsubsection{Authorship Attribution Challenge}
We evaluated the performance of our technique on the English-language part of the cross-domain authorship attribution challenge \citep{qi2018universal}. This challenge involves $4$ independent authorship attribution problems with $k$ candidate authors for $k\in \{5,\,10,\,15,\,20\}$. For each author in each problem, a corpus containing $7$ different labeled documents is provided. Each problem is also provided with a set of unlabeled documents. The goal is to correctly attribute the authorship of each document in the test set to one of the $k$ candidate authors in each problem.
\par
We used our HC-based approach to solve each problem by attributing each document from the test set to whichever author has the smallest index of discrepancy between this document and the corpus of that author in the training set. 
\newtext{
The vocabulary $W$ was formed for each specific problem using $3,000$ of the most common words, word bigrams, and word trigrams over all documents in the training set. Before counting word n-grams, we removed proper names and cardinal numbers and converted words to their dictionary form using the lemmatizer described in \citep{qi2018universal}.}
For the problems with 5, 10, 15, and 20, authors, our technique attained accuracies of $0.75$, $0.775$, $0.5$, and $0.43$, respectively. Our results imply an average F1 score of $0.75$ -- the second-best score reported for this part of the challenge; see \citep{kestemont2018overview}.

    
    


\section{
Analyzing Success in Authorship Attribution \label{sec:success}}
In this section, we suggest an explanation for the observed success of the HC-based test in discriminating between authors. We consider word counts under a variance-stabilizing transformation and analyze their variance across documents within a corpus of homogeneous authorship. 
We observe that the HC-discrepancy between a document and the corpus of an author is mostly affected by words characteristic of the author and not by words characteristic of topics in the text. 

\subsection{Author-Characteristic Words}
We propose that a word truly {\it characteristic} of an author would be used consistently across documents by that author. In contrast, a {\it topic-related} word will occur very frequently in documents associated with that topic, but not frequently in documents associated with unrelated topics. 
A simple model articulating this distinction says that words characteristic of an author are sampled independently from a multinomial distribution that is fixed across the corpus, whereas topic-related words are sampled via more structured mechanisms \citep{griffiths2004hierarchical, blei2007correlated, chang2010hierarchical, deng2014association, MargaretAiroldi2016,ross2019tracking}. 
%
Therefore, if $\mu(w)$ is the underlying frequency of the characteristic word $w$ in this multinomial distribution, the count of $w$ in the document $D$ is modeled by a Poisson distribution with parameter $\lambda(w|D) = \mu(w)|D|$ where $|D|$ denotes the total number of words in $D$. In contrast, the count of a topic-related word may follow a Poisson mixture distribution or may even be affected 
by stochastic dependence structures among words associated with the same topic \citep{blei2007correlated}. 
Such effects increase the variance of counts of topic-related words across documents within a corpus, resulting in overdispersion with respect to the Poisson sampling model \citep{breslow1984extra} \citep[Ch. 6.2.3]{mccullagh1989generalized}. 
In practice, the Poisson sampling model for words that are not topic-related does not match observed word counts \citep{mosteller2012applied, church1995poisson}. Nevertheless, a prediction of the model \-- relative variance as a measure for topic-relatedness \-- seems to hold in the cases we examined. 

\subsection{Variance-Stabilizing Transformation}
\begin{figure}
    \centering
    \includegraphics[scale =  0.13]{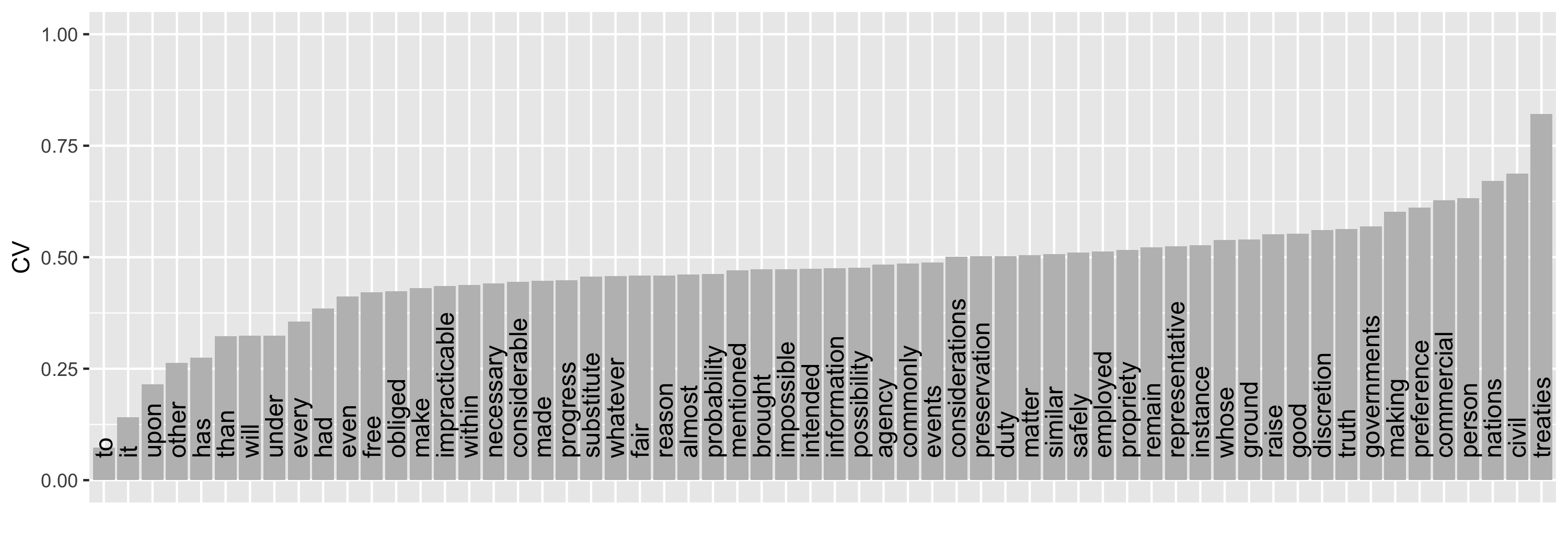}
    \caption{Examples of coefficient of variation $\CV(w|\Ccal)$ for selected words within a single corpus of homogeneous authorship $\Ccal$. We assume that words characteristic of the author mostly appear at the left-hand side of this plot. Proper names and cardinal numbers were removed.
    }
    \label{fig:null}
\end{figure}

We transform word counts using the transformation: \begin{align}
    \label{eq:Anscomb}
r(w|D) \equiv 2\sqrt{\frac{N(w|D)+\frac{1}{4}}{|D| }}.
\end{align}
\newtext{This version of the variance-stabilizing transformation is based on a suggestion in \citep{brown2001root}. If $N(w|D)$ follows a Poisson distribution with parameter $\mu(w)|D|$ where $|D|\gg \mu(w)$, then the distribution of $r(w|D)$ is approximately normal with mean $2\sqrt{\mu(w)}$ and variance $1/|D|$. By considering documents of roughly equal number of words within the same corpus, we assume that this variance is constant across documents belonging to a corpus.} 
Overdispersion with respect to the Poisson sampling model of a word $w$ implies that the variance of $r(w|D)$ across a corpus is larger than the naively-expected variance $1/|D|$. Therefore, we think of this variance as a measure of topic-relatedness of $w$. 
According to standard linear discriminant analysis principles, we consider a coefficient of variation (CV) based on the ratio of the squared root of this variance to the mean of $r(w|D)$.
Specifically, define the sample mean and variance across documents within a corpus $\Ccal$ as
\begin{align*}
    & \mu(w|\Ccal) \equiv \Ave\left(\{r(w|D) \}_{D \in \Ccal} \right),\\
    &  \sigma^2(w|\Ccal) \equiv \Var\left(\{r(w|D) \}_{D \in \Ccal} \right),
\end{align*}
respectively. We use the CV, defined as 
\[
CV(w|\Ccal) \equiv \frac{\sigma(w|\Ccal)}{ \mu(w|\Ccal)},
\]
as a measure for the variability of the word $w$ within the corpus $\Ccal$. 
Figure~\ref{fig:null} illustrates examples of $\CV(w|\Ccal)$ for randomly selected words from Hamilton's corpus in the Federalist Papers. 
The multinomial sampling model for author characteristic words predicts that such words typically appear on the left-hand side of Figure~\ref{fig:null}. In what follows, we verify that HC is mostly affected by such words. 

\subsection{Across-Corpus Coefficient of Variation versus P-value}
In order to identify words likely to influence the HC-discrepancy heavily, we perform many two-sample HC tests involving document-corpus pairs and quantify the properties of the words in the word-list selected by the HC calculation.
Given a document $D$, a corpus $\Ccal$ and a vocabulary $W$ of words, the P-values with respect to the binomial allocation model between the two word-frequency tables provide an ordering of the words in $W$. For each position in this ordering, we record $\CV(w|\Ccal)$ of the word $w$ appearing at that position and average the result over multiple document-corpus pairs. For a document-corpus pair in which the document happens to be a member of the corpus, we remove the document from the corpus before applying the HC test.
\begin{figure}
    \centering
    \includegraphics[scale = 0.16]{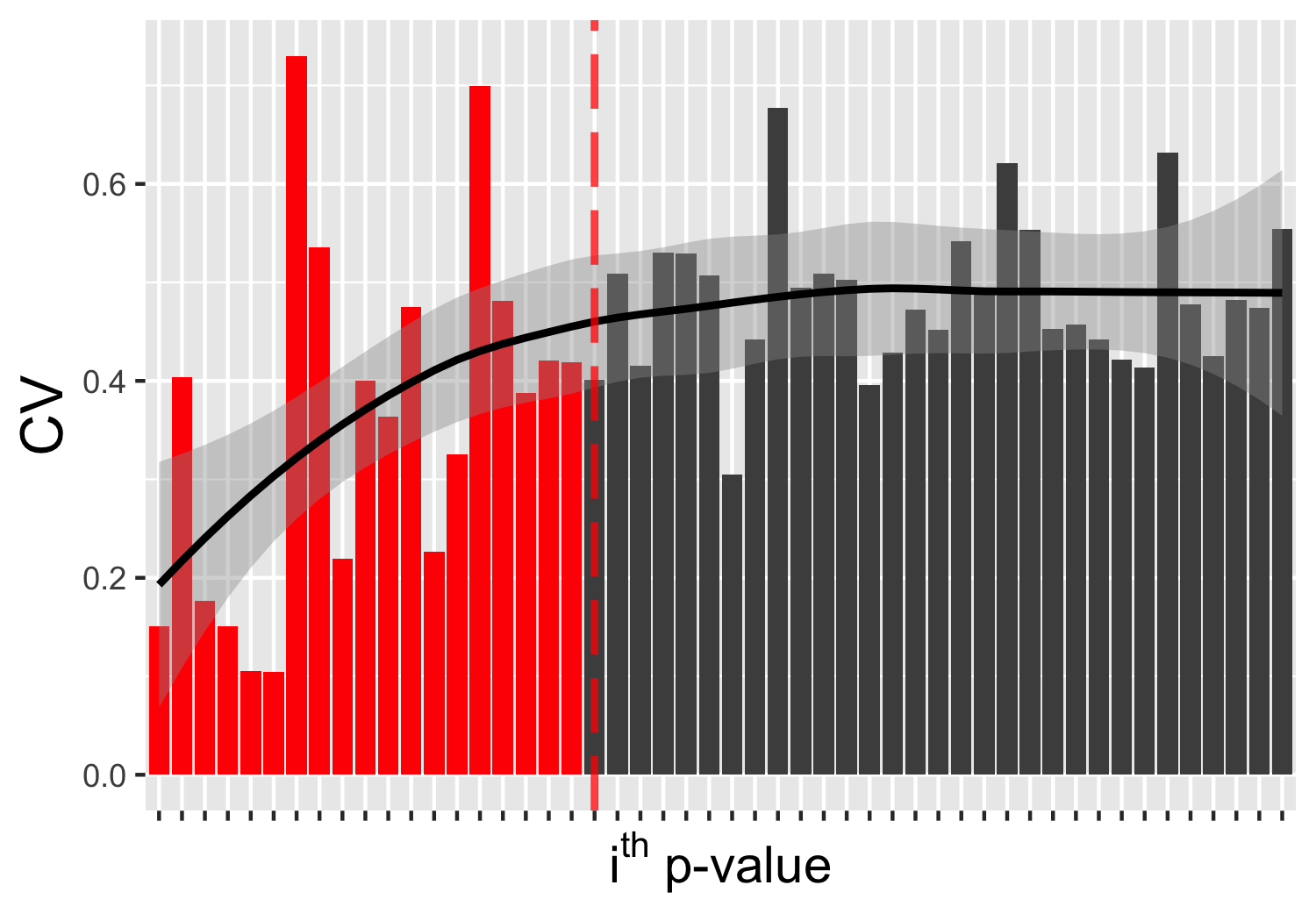}
    \includegraphics[scale = 0.16]{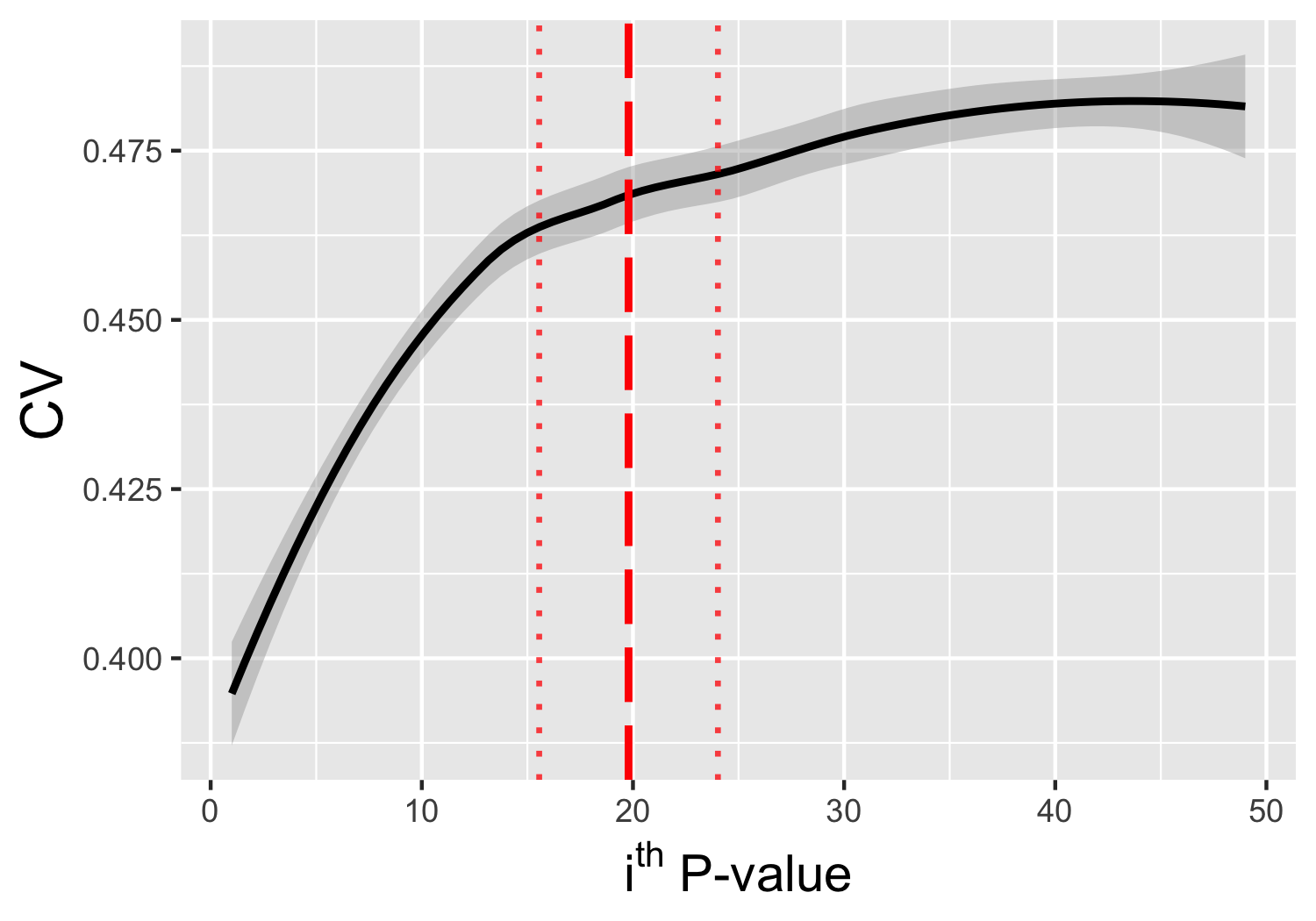}
    
    \includegraphics[scale = 0.16]{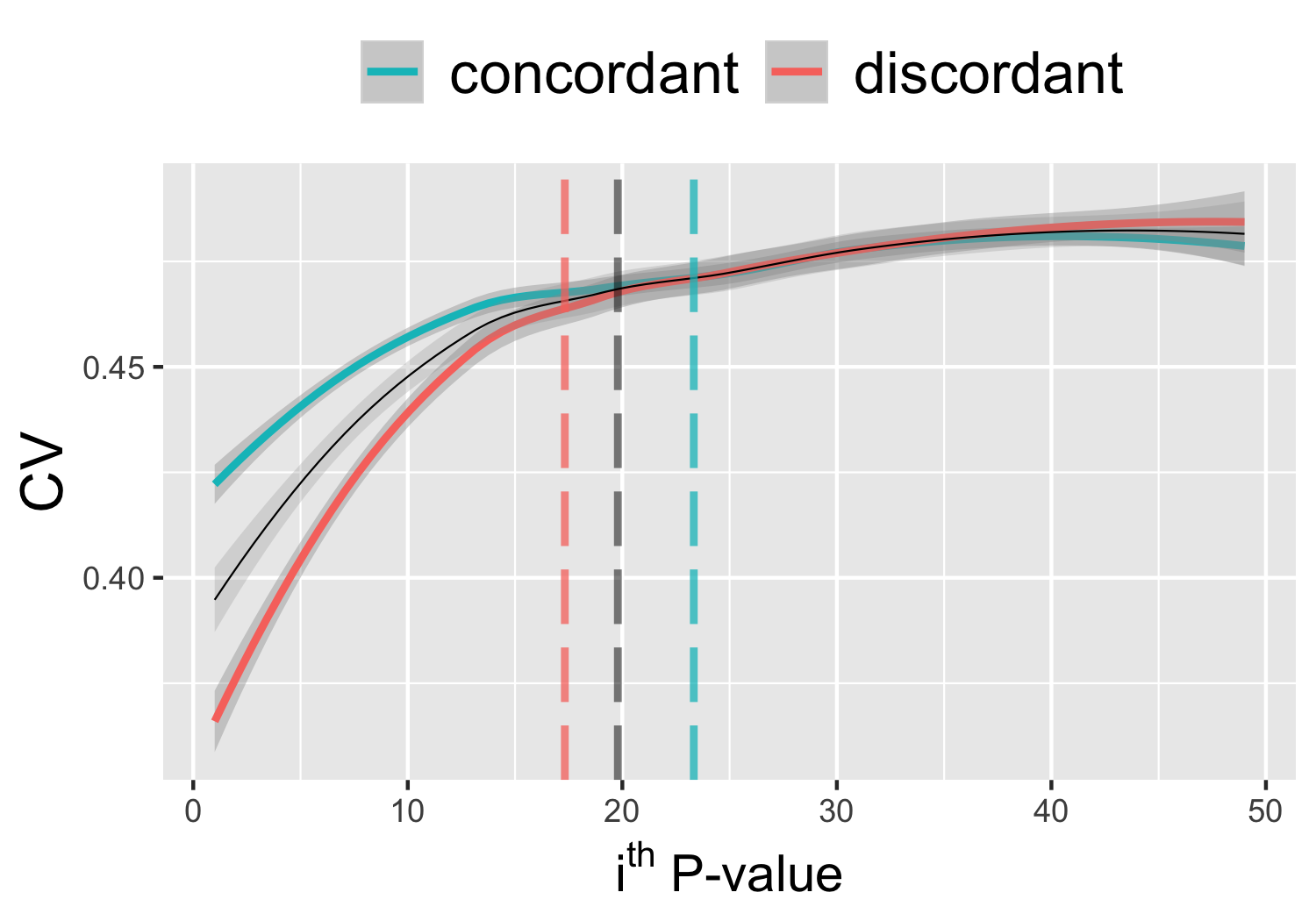}
    
    \caption{
    \newtext{
    Coefficient of variation $\CV(w|\Ccal)$ within a corpus ordered according to the rank of the P-value of the word $w$ in testing individual documents against a corpus $\Ccal$. Proper names and cardinal numbers were removed.
    Top: Results from a single test. Words falling below the HC threshold are indicated in red. The smooth line is a LOESS curve fit. Middle: Average of $\CV(w|\Ccal)$ for each rank over 1000 document-corpus pairs. The vertical line indicates the mean value of the HC threshold, while the vertical dashed lines indicate the range of the HC threshold in $95\%$ of the cases.
    Bottom: Average of $\CV(w|\Ccal)$ for each rank over 1000 document-corpus pairs while distinguishing cases when tested document is from the corpus of the same author (concordant) and the corpus of a different author (discordant). Vertical lines indicating the mean value of the HC threshold. The dark line is the global average, also given in the middle panel. 
    }}
    \label{fig:var_analysis1}
\end{figure}
Figure~\ref{fig:var_analysis1} illustrates the results of this evaluation: The top frame shows values of $\CV(w|\Ccal)$ ordered according to the P-value of $w$ obtained in a single test of one document against the corpus $\Ccal$. The middle frame reveals the trend seen in the top frame by showing the average of $\CV(w|\Ccal)$ at each location across multiple document-corpus pairs. 
If follows that, on average, a word $w$ associated with a small P-value is also associated with a small $\CV(w|\Ccal)$. This CV serves as a measure of the degree to which a word is author-characteristic versus topic-characteristic. We conclude that {\it words associated with small P-values are typically author-characteristic}, suggesting that the HC-discrepancy is mostly affected by author-characteristic words, and explaining to some degree why it discriminates well between documents and corpora of different authorship.
%

    

\subsection{Concordant and Discordant Tests}
As a final illustration for our proposed interpretation of factors driving the success of our HC-based test in authorship attribution, we distinguish between the case where the document and corpus in each test have the same author (concordant) or not (discordant). Namely, we repeat the testing and averaging procedure outlined above, but, in addition, we mark whether the document and corpus are concordant or discordant. The results of this procedure are illustrated in the bottom panel of Figure~\ref{fig:var_analysis1}. This figure shows that the averaged CV is significantly smaller in discordant pairs and that the HC threshold appears to derive the change between the curves. Since smaller P-values means larger HC, this situation is in agreement with the observation that HC is affected by P-values of words associated with smaller CVs across a corpus of homogeneous authorship. 

\section{Conclusions}
\label{sec:cocnlusion}
We developed a technique to measure the similarity/discrepancy of two word-frequency tables and applied it to authorship attribution. Our measure, $\dist_{\HC}$, uses a word-level binomial allocation model to form word-by-word P-values, which are then combined using the HC statistic. The HC calculation also identifies a set of words where there seem to be notable differences between two tables. \par
When applied to authorship attribution challenges, we measure the value of the novel document's $\dist_{\HC}$ score relative to each corpus, attributing authorship based on the smallest rank-score. This automated procedure gives results comparable to previous studies, but without handcrafting or tuning. 
In analyzing the ingredients for the success of our technique in authorship attribution, we found that, in practice, our discrepancy measure is mostly affected by words associated with low variance within a corpus of homogeneous authorship. 

\begin{supplement}
\stitle{HCAuthorship}
\sdescription{
The code and data for generating all figures and tables in this paper. 
}
\end{supplement}

\section{Acknowledgments} 
The author would like to thank David Donoho for fruitful discussions and to three anonymous reviewers for providing comments that have greatly improved this paper.

\bibliography{HigherCriticism}

\end{document}

%% file: macros.tex
\newcommand{\numOfFedWords}{$378$ }

\newcommand*{\CV}{\mathrm{CV}}

\newcommand*{\Var}{\mathrm{Var}}
\newcommand*{\Ave}{\mathrm{Ave}}
\newcommand*{\HC}{\mathrm{HC}}
\newcommand*{\PD}{\mathrm{d_{\lambda}}}
\newcommand*{\Ham}{\mathrm{Hamilton}}
\newcommand*{\Mad}{\mathrm{Madison}}

\newcommand*{\Ccal}{\mathcal{C}}

\newcommand{\Bin}{\mathrm{Bin}}
\newcommand{\Prob}{\mathrm{Prob}}

\newcommand*{\argmin}{\mathrm{argmin}}
\newcommand*{\argmax}{\mathrm{argmax}}

\newcommand{\newtext}[1]{{\textcolor{black}{#1}}}

\usepackage{enumerate}

\newcommand{\dist}{\mathrm{d}}
\newcommand{\cossim}{\mathrm{d_{\cos}}}

\def\spacingset#1{\renewcommand{\baselinestretch}%
{#1}\small\normalsize} \spacingset{1}

%% file: table_results.tex
\begin{tabular}{p{27mm}|c|c|c}
    vocabulary size & $250$ & $1,000$ & $3,000$ \\
    \hline
    $\HC^\dagger$ &  {\bf .791} (.0092) & {\bf .833} (.0093) & {\bf .851} (.0097)\\
    \hline
    $\HC^*$ &  {\bf .787} (.0063) & { \bf .831} (.0074) & { .840} (.0103) \\
    \hline
    $\PD_{0}$ ($G^2$) & .724 (.0158) & .794 (.0010) & .829 (.0121) \\ 
    \hline
    $\PD_{2/3}$ (Cressie-Reed) & .723 (.0115) & .774 (.0088) & .780 (.0179) \\ 
    \hline
    $\PD_2$ (Pearson's $\chi^2$) & .718 (.0142) &  .747 (.0130)  & .717 (.0117) \\
\hline 
cosine discrepancy & .529 (.00096) & .553 (.0145) & .571 (.0120)
\\ \hline
5-nearest neighbors
cosine discrepancy &  .697 (.0147)  & .712 (.0084)  & .722 (.0122)
\end{tabular}
